\pdfoutput=1

\documentclass[11pt]{article}

\usepackage{EMNLP2022}

\usepackage{times}
\usepackage{latexsym}

\usepackage[T1]{fontenc}

\usepackage[utf8]{inputenc}

\usepackage{microtype}

\usepackage{inconsolata}

\usepackage{helvet}  
\usepackage{courier}  
\usepackage{url}  
\usepackage{xurl}
\usepackage{graphicx}  
\usepackage{CJKutf8}
\usepackage{latexsym}
\usepackage{multirow}
\usepackage{amsmath}
\usepackage{booktabs}
\usepackage{subfigure}
\usepackage{pdfpages}
\usepackage{array}
\usepackage{makecell}

\providecommand{\tightlist}{%
      \setlength{\itemsep}{0pt}\setlength{\parskip}{0pt}}
\usepackage{enumitem}
\title{CrossDial: An Entertaining Dialogue Dataset of Chinese Crosstalk}

\author{Baizhou Huang$^{1,2}$, Shikang Du$^3$ \and Xiaojun Wan$^{1,2}$ \\
        $^1$Wangxuan Institute of Computer Technology, Peking University\\
        $^2$The MOE Key Laboratory of Computational Linguistics, Peking University\\
        $^3$Ecole Polytechnique\\
        \texttt{\{hbz19,wanxiaojun\}@pku.edu.cn}\\
        \texttt{shikang.du@polytechnique.edu}}

\begin{document}
\begin{CJK*}{UTF8}{gbsn}
\maketitle
\begin{abstract}
	Crosstalk is a traditional Chinese theatrical performance art. It is commonly performed by two performers in the form of a dialogue. With the typical features of dialogues, crosstalks are also designed to be hilarious for the purpose of amusing the audience. In this study, we introduce \textbf{CrossDial}, the first open-source dataset containing most classic Chinese crosstalks crawled from the Web. Moreover, we define two new tasks, provide two benchmarks, and investigate the ability of current dialogue generation models in the field of crosstalk generation. The experiment results and case studies demonstrate that crosstalk generation is challenging for straightforward methods and remains an interesting topic for future works.
	
\end{abstract}

\section{Introduction}
Crosstalk, also known by its Chinese name \textit{相声/xiangsheng}, is a traditional Chinese theatrical performance art. It is commonly performed by two performers. One performer is the leading role ({\it 逗哏/dougen} in Chinese) and the other is the supporting role ({\it 捧哏/penggen} in Chinese). 

\begin{figure}[htb]
    \centering
    \includegraphics[width=2.9in]{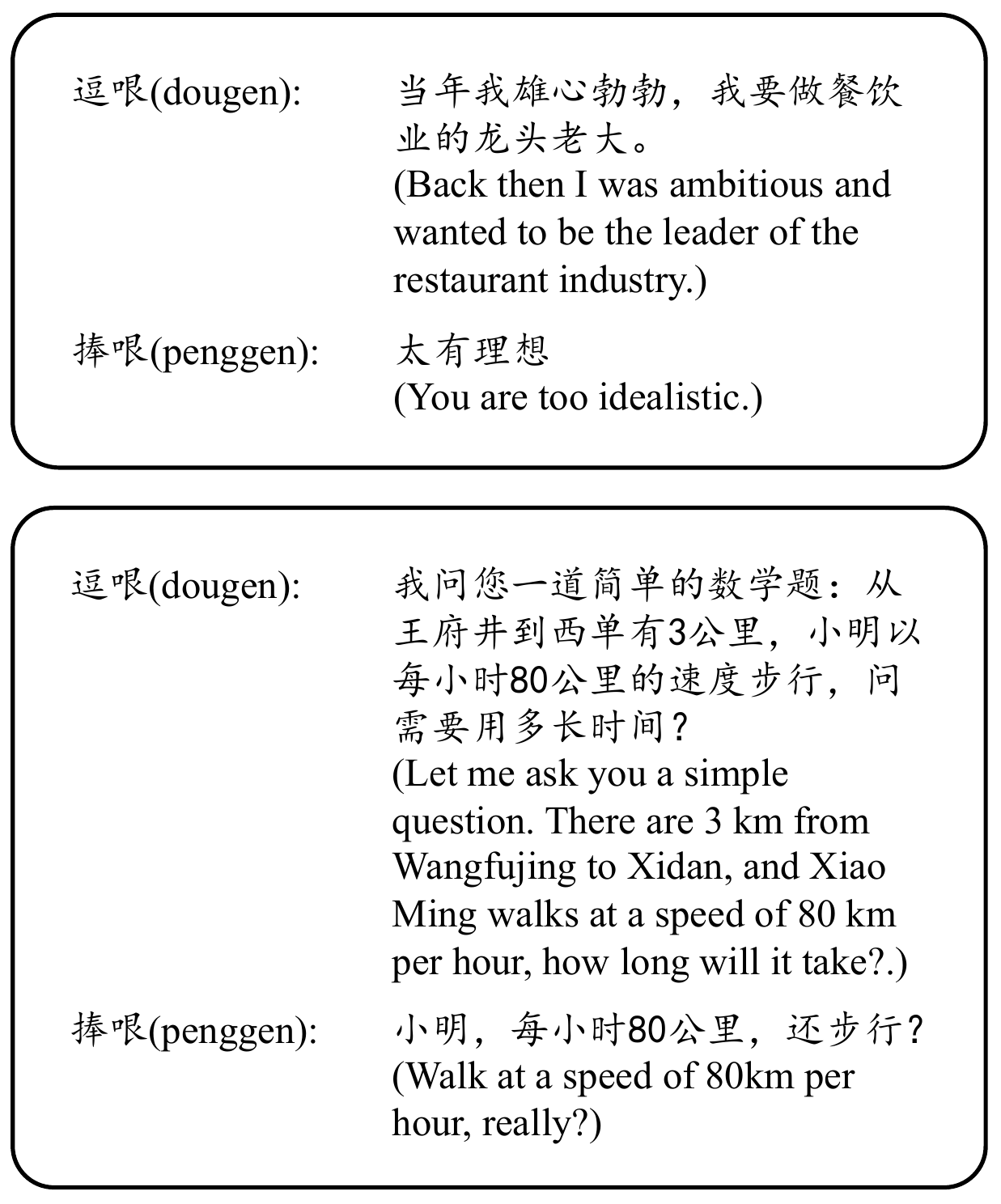}
    \caption{Two excerpts of Chinese crosstalk. In the first one, dougen starts the topic about his dream, while penggen comments with negative attitude. In the second one, dougen asks penggen a mathematical problem, while penggen unexpectedly captures the nonsense in the problem that one person walks at a speed of 80km per hour.}
    \label{fig:intr_example}
\end{figure}

The form of crosstalks is just like chat or gossip with two persons responding to each other alternately. But there are several conventional performance patterns in crosstalks that are different from daily dialogues. First, the crosstalk is a complete story with one main topic to entertain the audience. The two performers should discuss strictly around the main topic instead of changing topics casually like gossip. Second, the language patterns of the two performers are different in crosstalks. The leading role is the one who dominates the dialogue and drives the plot forward. Mostly, the leading role tells stories and jokes during the performance. On the other hand, the supporting role gives short comments to support or question the leading role's opinion.  In some cases, the supporting role may point out the humorous point in the leading role's utterance to explain to the audience, or even add fuel to the fire to make it funnier. Third, the crosstalk language is rich in comedy acting skills, such as puns, and is usually delivered in a rapid, bantering style. For the purpose of bringing laughter to the audience, the language of crosstalk features humorous dialogues \cite{link1979genie,moser1990reflexivity,mcdougall1984popular}. We provide two excerpts in Figure \ref{fig:intr_example}.

In the study, we are concerned with generating Chinese crosstalks automatically. Currently, the traditional art is suffering from the lack of scripts which is hard to write even for humans. It is of high artistic value to design a model that can automatically generate crosstalks. Apart from this, the ability to generate entertaining dialogue responses is also very useful in many commercial products (e.g. chat-bots) by making them more appealing. 
Though daily dialogue generation has been widely explored and achieved great success in previous studies \cite{li-etal-2016-deep,li-etal-2017-adversarial,wen-etal-2017-network},
it remains unknown whether entertaining dialogues can be automatically generated or not. 
The special language style and the two-role pattern of crosstalks make it a challenging but interesting task to be explored. 

To support research on automatic crosstalk generation, we build \textbf{CrossDial}, the first open-source crosstalk dataset that covers most classic Chinese crosstalks. It is a large-scale dialogue dataset consisting of 1257 crosstalk scripts and 140432 data samples crawled from the Internet. 

To investigate the automatic generation of such entertaining dialogues over the proposed dataset, we design two different tasks. The first is a generation task, i.e. \textbf{Crosstalk Response Generation}. That is, given several continuous utterances as context, the model is required to generate the next utterance as response. Considering the one-to-many problems in current generation metrics and the difficulty in automatic humor evaluation, we then additionally introduce a more basic retrieval task, i.e. \textbf{Crosstalk Response Selection}. That is, given several continuous utterances as context, the model is required to find the best response from the supported choices. 

We implemented several typical neural models as baselines and evaluate them on the newly defined tasks. The results of automated metrics show the difficulty of our proposed tasks. The human evaluation and case studies further demonstrate the challenges of generating crosstalks. 

The contributions of this paper are summarized as follows:
\begin{enumerate}[label={\arabic*})]
\tightlist
\item We propose the first open-source Chinese crosstalk dataset which contains most classic Chinese crosstalks. The dataset will be released. 
\item We design two different tasks and provide two benchmarks respectively for mainstream methods. 
\item Both automatic evaluation and human evaluation are performed to evaluate the ability of typical models for automatic crosstalk generation.
\end{enumerate}

\section{Related Work}
\subsection{Dialogue Dataset}
With the explosive growth of social networks, a large number of dialogue corpora have been collected from multiple data sources\cite{danescu-niculescu-mizil-lee-2011-chameleons,tiedemann-2012-parallel,lowe-etal-2015-ubuntu,wu-etal-2017-sequential}. For example, {\bf Cornell Movie-Dialogue Corpus}\cite{danescu-niculescu-mizil-lee-2011-chameleons} and {\bf OpenSubtitles}\cite{tiedemann-2012-parallel} collected dialogues from scripts of movies and TV series. The {\bf Ubuntu Dialogue Corpus}\cite{lowe-etal-2015-ubuntu}
 and {\bf Douban}\cite{wu-etal-2017-sequential} collected unstructured dialogues from large-scale comments on social media. {\bf Persona-Chat}\cite{zhang-etal-2018-personalizing} collected dialogues where each participant plays the part of a specific persona. The area of dialogue generation has witnessed great developments based on these resource-based studies. 
 
\subsection{Dialogue Generation Architectures}
As we mentioned above, Chinese crosstalk is a special form of dialogue. It is trivial to generate crosstalks similarly to dialogue generation. Previous works in this field relied on rule-based methods, from learning generation rules from a set of authored labels \cite{oh-rudnicky-2000-stochastic}
to building statistical models based on templates and heuristic rules \cite{817450}.  
After that, information retrieval (IR) based methods \cite{https://doi.org/10.48550/arxiv.1408.6988} and the statistical machine translation (SMT) based methods \cite{ritter-etal-2011-data} dominated this field gradually.

More recently, the neural network was introduced in this field in the form of end-to-end learning methods \cite{https://doi.org/10.48550/arxiv.1507.04808,li-etal-2016-deep,li-etal-2017-adversarial}. Among numerous neural models, transformer\cite{transformer} has proven to be one of the best backbone models and has been applied to various tasks. On the basis of that, several pretrained language models\cite{devlin-etal-2019-bert, Radford2018ImprovingLU} were proposed and the pretrain-finetune paradigm has achieved great success. Therefore, we chose transformer-based pretrained language models as baseline models.

\subsection{Computational Humor}
Computational humor is also related to this study. Humor detection and humor generation are two main topics in this field.
Humor detection is commonly formalized as a classification task. Plenty of methods have been applied to solve this problem\cite{yang-etal-2015-humor,chen-soo-2018-humor,weller-seppi-2019-humor}. Humor generation is much more challenging than humor detection. Previous studies mainly focused on one specific type of humor, such as puns\cite{yu-etal-2018-neural,he-etal-2019-pun,yu-etal-2020-homophonic}. The crosstalk is a comprehensive performing art consisting of many forms of humor, such as homophone, hyperbole, sarcasm and so on. Although we only conducted experiments over straightforward methods, we believed it would be beneficial to introduce semantic structure of humor into the models. Particularly, \citet{https://doi.org/10.48550/arxiv.1711.00294} preliminarily discussed the automatic generation of crosstalks. But it mainly focused on SMT based methods and the dataset is not released to the community. In this study, we provided benchmarks for most neural network-based methods, especially pretrained language models. Moreover, we carefully collected and cleaned the crosstalks scripts from the Internet, constructed the first open-source crosstalk dataset with both response generation and response selection tasks.

\section{Task Definition}
We formulate the problem of crosstalk generation as the next utterance prediction task as in daily dialogue generation. In particular, we define two sub-tasks namely Crosstalk Response Generation (CRG) and Crosstalk Response Selection (CRS). Given continuous utterances as context $c = \{u_1, u_2, ..., u_{n-1}\}$, the agent is required to generate the next utterance as response $r = u_n$ in the CRG task or distinguish the positive response $x^{pos} = u_n$ from the other three distractors $\{x^{neg}_0, x^{neg}_1, x^{neg}_2\}$ in the CRS task.

In the CRG task, the generated response is expected to be grammatical and coherent to the context, as in the general dialogue generation task. It should also be compatible with the specific pattern of the role that the agent plays. Moreover, the level of amusement and humor should be taken into consideration.

CRG is a one-to-many problem. In other words, there are many responses appropriate for one given context. Most of the current generation metrics (e.g. BLEU) are based on the comparison between the reference and the generated response. Therefore they cannot reflect the true level of the agent's generation ability. As a complement,
we introduce the CRS task to evaluate the agent's capability more objectively.

\section{CrossDial Dataset}
\subsection{Overview}
\begin{table*}[htb]
    \centering
    \begin{tabular}{c|c|c|c}
        \toprule
        Subsets & Sample Num(trian/valid/test) & Context Len(avg/max/min) & Response Len(avg/max/min)\\
        \midrule
        dougen & 75944 / 8628 / 4664 & 302.80 / 3460 / 11 & 22.01 / 127 / 2 \\
        \midrule
        penggen & 43372 / 5136 / 2688  & 319.20 / 3389 / 20 & 12.98 / 126 / 2 \\
        \bottomrule
    \end{tabular}
    \caption{Basic statistics of \textbf{CrossDial}. The lengths of context and response are measured in characters.}
    \label{tab:statistics_sample}
\end{table*} 
The objective of this work is to introduce the task of crosstalk generation and facilitate the study of both the CRG and CRS tasks. For this, we propose a large-scale Chinese crosstalk dataset \textbf{CrossDial}, a web-crawled dataset that covers most classic Chinese crosstalks. The dataset contains two types of utterances, i.e., \textbf{dougen} and \textbf{penggen}, corresponding to the leading role and the supporting role. Each sample for the CRG task consists of two fields: {\it context} and {\it positive response}, and each sample for the CRS task involves with three additional {\it negative responses (distractors)} constructed by us.

The dataset creation consists of three stages:

1) Data Collection: we crawled a set of $1257$ crosstalk scripts from the Internet which contains most Chinese crosstalks. 

2) Sample Creation: we split all crosstalk scripts into context-response pairs as data samples for the CRG task. To be compatible with the two-role patterns in crosstalk, we divided the dataset into two subsets.

3) Distractor Generation: we designed delicate distractors for the CRS task. To avoid false negatives of distractors, we recruited eight annotators to review all the distractors and filtered invalid ones.

After all, we created a dataset consisting of 140432 data samples in the form of context-response pair. Basic statistics of \textbf{Crossdial} are shown in Table \ref{tab:statistics_sample}. In the following, we will describe all stages in more detail.

\subsection{Data Collection}
We crawled a total of $1,551$ excerpts of classic crosstalks scripts from the Internet \footnote{\url{http://www.xiangsheng.org},  \url{http://www.tquyi.com}, et al.}. Due to reproductions among websites, one crosstalk script might be collected from different sources. Therefore, we only kept one script and dropped the other copies. In detail, two scripts were considered the same if they have an overlap of more than 15 seven-word-longer utterances\footnote{The thresholds were set based on manual inspection of the
data}. We also noticed several similar crosstalks because of a large number of script adaptations. We kept them as the status quo since they were indeed different scripts. We also took several heuristic methods for data cleaning. Finally, a total of $1257$ crosstalks were collected after this process.

\subsection{Sample Creation}
We extracted continuous utterances from collected crosstalk scripts as context-response pairs. Specifically, for each utterance in crosstalk scripts noted as {\it positive response}, we extracted the sequence of no more than twenty utterances prior to it as {\it context}.

With the above extraction process, the response utterance in one sample may appear in the contexts of others. To avoid information leakage from the test set to the training set, we split train, validation, and test sets at the granularity of scripts instead of context-response pairs. To be exact, we randomly sampled 75 scripts for the test set, 175 scripts for the validation set, and 1007 scripts for the training set.

Considering the different speech patterns between the leading role and the supporting role, it is interesting to divide the dataset into two subsets: \textbf{dougen} and \textbf{penggen}. Each subset included only the context-response pairs where the response belonged to the corresponding role. 

For the fast-paced performance before an audience, many utterances in crosstalks are designed to be short and meaningless, especially for the supporting role's lines. This is called \textit{generic response} \cite{li-etal-2016-diversity} in NLG, which may impede the diversity of dialogue systems. So we created a set of common and meaningless words. We removed data samples of which over half words of the response were in the set. We also limited the text lengths of responses to [2, 128) to avoid too long responses.

\subsection{Distractor Generation}
We generated distractors for the CRS task. It is both time-consuming and expensive to crowd-source human-written distractors for such a large dataset. Mostly, distractors are sampled randomly in previous work \cite{zhang-etal-2018-personalizing, lowe-etal-2015-ubuntu, wu-etal-2017-sequential}. We argue that randomly sampled distractors are so simple that models may leverage shortcuts to achieve better performance. For example, random distractors commonly have fewer n-gram overlaps with context than golden response. Instead, we aim to generate high-quality distractors that are 1) similar to the golden response in order to avoid the model from using shortcuts, and 2) consistent with the semantics of context to confound the model. 

We proposed two similarity-based methods to retrieve distractors from the dataset that satisfy the 
above two requirements. We used the cosine distance of sentence embeddings to measure the similarity between two utterances. Pretrained language models have achieved state-of-the-art performance in the field of sentence embeddings \cite{reimers-gurevych-2019-sentence, gao-etal-2021-simcse, kim-etal-2021-self, giorgi-etal-2021-declutr}. Given this fact, we borrowed the off-the-shelf tool, sentence transformers\footnote{\url{https://github.com/UKPLab/sentence-transformers} and used two pretrained models \href{https://huggingface.co/uer/sbert-base-chinese-nli}{sbert-base-chinese-nli} and \href{https://huggingface.co/sentence-transformers/distiluse-base-multilingual-cased-v1}{distiluse-base-multilingual-cased-v1}.} to generate sentence embeddings for all utterances. 

\textbf{Response-Similar Distractor} For every data sample, we searched the corpus for similar responses as distractors with the golden response as the query. If the similarity score of the two responses is high, we consider the extracted one as a high-quality distractor. 

\textbf{Context-Consistent Distractor} For every data sample, we searched the corpus for similar contexts with the context of the current sample as the query and took the corresponding response as the distractor. We regard the chosen distractor to be consistent with the current context since it is the golden response to the searched context, and the two contexts resemble each other in semantics. An example of the generated distractors is shown in Table \ref{tab:distractor}.

\begin{table}[htb]
    \small
    \centering
    \begin{tabular}{m{2cm}<{\centering}|p{5cm}}
        \toprule
        Context & 您还喜欢看小说？\newline
                Do you also like fictions?\\
        \midrule
        Response & 不是看而是研究，尤其是对我国古典小说像《列国》、《水浒》、《红楼》、《西游》我都爱看，特别是对《三国演义》我敢说有独特的见解。\newline
        Do research instead of just reading. I'm quite a fan of Chinese classical fictions such as {\it lieguo}, {\it shuihu}, {\it honglou}, {\it xiyou}. Especially for {\it sanguo}, I have unique insights about it.\\
        \midrule
        RSD & 你要说三国，水浒古典名著那您研究研究，这个有点意思。\newline
        It's worth studying if you're talking about classical fictions such as {\it sanguo} and {\it shuihu}. These are quite interesting.\\
        \midrule
        CCD & 噢，我喜欢看的那都是古典文学呀。\newline
            Oh, all I'd like to read is classical literature.
            \\
        \bottomrule
    \end{tabular}
    \caption{An example of generated distractors. RSD is short for response similar distractor and CCD is short for context consistent distractor.}
    \label{tab:distractor}
\end{table} 

Extremely confusing and disorienting though the generated distractors are, it is worth considering that they can be appropriate responses. As seen in Table \ref{tab:distractor}, the CCD can also be used as response to the context. To avoid the false negatives of distractors, we set a threshold $\lambda$ \footnote{$\lambda$ was tuning according to manual inspection of samples. Finally, we set $\lambda = 0.8$ .} to filter all retrieved outputs that had too high cosine similarity. 

Apart from automatic filtering rules, we randomly drew two percent of generated distractors for quality evaluation. We found that only 5.28\% of distractors were false negative. It proved that it was feasible to use generated distractors as negative responses.

In particular, for the test set, we recruited eight expert annotators to review all distractors. To simulate the process in which a model generates response conditioning on the context, we provided annotators with six rounds of prior utterances as context. Annotators were asked to select all appropriate responses from a supported choice set that is composed of generated distractors. We also added the golden response to the choice set to ensure the quality of annotation results. After annotation, we dropped invalid distractors, and randomly sampled three negative responses out of the rest for each sample.

\section{Experiment}

We implemented several typical models, experimented on our proposed CrossDial dataset, and provided performance benchmarks for both CRG and CRS tasks. In the following, we first introduce the automated metrics used for evaluation. Then we report the results of commonly used models. At last, we show the human evaluation and case studies of the generated responses.

\subsection{Automated Metrics}
\begin{table*}[htb]
    \centering
    \scalebox{1}{
    \begin{tabular}{c|p{2cm}<{\centering}p{2cm}<{\centering}p{2cm}<{\centering}p{2cm}<{\centering}p{2cm}<{\centering}p{2cm}<{\centering}}
        \toprule
        Method & perplexity & BLEU-4 & ROUGE-2 & ROUGE-L & Distinct-1 & Distinct-2 \\
        \midrule
        \multicolumn{7}{c}{penggen} \\
        \midrule
        \textbf{Trans} & 14.63 & 3.64 & 1.59 & 15.17 & 1.37 & 6.14 \\
        \textbf{BART} & 7.76 & 5.55 & 6.41 & 19.61 & 7.37 & 37.95\\
        \textbf{GPT} & 8.09 & 3.49 & 4.18 & 19.80 & 4.74 & 22.79\\
        \textbf{T5} & 8.80 & 5.75 & 6.71 & 21.75 & 5.47 & 32.54 \\
        \midrule
        \multicolumn{7}{c}{dougen} \\
        \midrule
        \textbf{Trans} & 15.30 & 2.21 & 2.06 & 15.43 & 1.37 & 9.39 \\
        \textbf{BART} & 9.41 & 2.98 & 4.19 & 18.00 & 3.69 & 29.16 \\
        \textbf{GPT} & 8.75 & 2.24 & 2.35 & 16.25 & 3.00 & 22.30 \\
        \textbf{T5} & 9.71 & 3.32 & 5.10 & 19.79 & 2.73 & 24.24 \\
        \bottomrule
    \end{tabular}}
    \caption{Comparison of generative models on the CRG task}
    \label{tab:crg_result}
\end{table*} 
We adopt perplexity, BLEU-4, ROUGE-2, ROUGE-L, Distinct-1, and Distinct-2 as automated metrics for the CRG task.
BLEU \cite{papineni-etal-2002-bleu} and ROUGE \cite{lin-2004-rouge} are the most commonly used metrics for natural language generating tasks. They are both based on the n-gram overlaps between a generated response and a reference response. Distinct-1 and Distinct-2 \cite{li-etal-2016-diversity} are also n-gram-based metrics and are used to measure the diversity of the agent's generated responses. We adopt accuracy for the CRS task since it is essentially a selection problem. 

\subsection{Baseline Methods}
We leveraged the open-source community huggingface\footnote{\url{https://huggingface.co/}} to build two classes of models: generative models and retrieval models. For the CRG task, we only experimented with generative models. Whereas for the CRS task, both classes were evaluated. Retrieval models take the concatenation of the context and one candidate response as input, and score it. Generative models take the context as input, and score each candidate with its generation probability. Both classes select the highest-scored candidate as the response.

In particular, we consider the following generative baselines:
\begin{itemize}
    \item {\sc \bf Trans}: Transformer \cite{transformer} is a \textsc{SEQ2SEQ} model which has been widely used in natural language processing. We adopted a Transformer-base model in the experiment.
    \item {\sc \bf BART}: BART \cite{lewis-etal-2020-bart} is a Transformer pretrained as a denoising autoencoder. We adopted BART-base\footnote{\url{https://huggingface.co/uer/bart-base-chinese-cluecorpussmall}} in the experiment.
    \item {\sc \bf GPT}\cite{Radford2018ImprovingLU}: GPT is also a pretrained language model. But unlike BART, it only has a Transformer decoder as backbone. We adopted GPT-2\footnote{\url{https://huggingface.co/uer/gpt2-chinese-cluecorpussmall}}.
    \item {\sc \bf T5}: T5 \cite{T5} introduced a unified framework that converts every natural language processing task into a text-to-text format. We adopted T5-base\footnote{\url{https://huggingface.co/uer/t5-base-chinese-cluecorpussmall}}.
\end{itemize}

We also consider the following retrieval baselines:
\begin{itemize}
	\item {\sc \bf Trans}$_{encoder}$: The encoder of the Transformer model. We adopted the Transformer-base encoder.
	\item {\sc \bf BERT}: BERT \cite{devlin-etal-2019-bert} is a pretrained Transformer encoder. \citet{BERT-wwm} proposed Whole Word Masking (WWM) to develop performance of original BERT in processing Chinese texts. We adopted BERT-wwm-base\footnote{\url{https://huggingface.co/hfl/chinese-bert-wwm-ext}}.
	\item {\sc \bf RoBERTa}: RoBERTa \cite{RoBERTa} is built upon BERT with modified key hyper-parameters. We adopted RoBERTa-wwm-base\footnote{\url{https://huggingface.co/hfl/chinese-roberta-wwm-ext}} in the experiment. The Whole Word Masking technique is also utilized.
	\item {\sc \bf ERNIE}:  ERNIE \cite{ERNIE} incorporates knowledge masking strategies to learn better language representation. We adopted ERNIE-1.0 \footnote{\url{https://huggingface.co/nghuyong/ernie-1.0}}.
\end{itemize}

We performed six hyper-parameters search trials for each model. Hyper-parameters and final checkpoints of baselines were both tuned with perplexity (for generative models) or accuracy (for retrieval models) on the validation set. 

\subsection{Result and Analysis}
\begin{table}[htb]
    \centering
    \scalebox{1}{
    \begin{tabular}{c|p{2cm}<{\centering}p{2cm}<{\centering}}
        \toprule
        Subset & penggen & dougen \\
        \midrule
        \multicolumn{3}{c}{Generative} \\
        \midrule
        \textbf{Trans} & 14.47 & 16.01\\
        \textbf{BART} & 38.61 & 33.16\\
        \textbf{GPT} & 42.11 & 38.59\\
        \textbf{T5} & 36.53 & 35.29\\
        \midrule
        \multicolumn{3}{c}{Retrieval} \\
        \midrule
    \textbf{Trans}$_{encoder}$ & 39.88 & 50.75  \\
        \textbf{BERT} & 74.21 & 81.47 \\
        \textbf{RoBERTa} & 76.19 & 79.60 \\
        \textbf{ERNIE} & 71.01 & 84.54 \\
        \textbf{SIM} & 30.02 & 29.69 \\
        \textbf{CLS} & 35.45 & 40.71\\
        \bottomrule
    \end{tabular}}
    \caption{Accuracy of various models on the CRS task. \textbf{SIM} and \textbf{CLS} are trivial methods for analysis.}
    \label{tab:crs_result}
\end{table} 

We trained and tested the baselines on the \textbf{penggen} and \textbf{dougen} subsets separately. The main results are reported in Table \ref{tab:crg_result} and Table \ref{tab:crs_result}.

\textbf{Pretrained or not?} We observe that pretrained language models outperform models without pretraining on both tasks, and the results proved the effectiveness of pretraining. However, the BLEU and ROUGE scores for the CRG task are still very low, showing the great challenge of response generation for crosstalks.

\textbf{Penggen vs. Dougen} Dougen is the one who drives the dialogue forward, while penggen often acts as a go-between with short sentences. Intuitively, the language patterns of dougen are more complex to learn. The results on the CRG task show that the same model performs worse in the \textbf{dougen} subset than in the \textbf{penggen} subset, which is consistent with this intuition. However, experiments on the CRS task show the opposite result that retrieval models perform worse on the \textbf{penggen} subset. We attribute the phenomenon to the different difficulties of distractors of the two subsets. Since the supporting role has a relatively fixed form of response, it is more likely to retrieve high-quality distractors for the \textbf{penggen} subset, which makes it harder than the \textbf{dougen} subset.

\textbf{Generative vs. Retrieval} Both generative models and retrieval models are able to handle the CRS task. The results indicate that retrieval models perform much better than generative models. It is obvious since the training objective of retrieval models is consistent with the CRS task. But still some generative models perform quite poor, especially \textbf{Trans}. The reason may be that the training of generative models is based on token-level loss while the CRS task requires a good measure of sentence-level probability to select the true response.

\textbf{Shortcut Analysis} Models might use unseen patterns as shortcuts in the CRS task. We proceeded with two trivial methods, \textbf{SIM} and \textbf{CLS}, to empirically negate the phenomenon in the proposed dataset. \textbf{SIM} is an untrained method that utilizes RoBERTa to acquire sentence embeddings for the context and all candidates, and scores the candidates with the cosine similarity. \textbf{CLS} is a trained selector that takes only the candidates without context as input. We adopted the Transformer-base encoder as its backbone model. The poor performance of both methods showed that the phenomenon is not notable in the proposed dataset.

\subsection{Human Evaluation}

\begin{table}[htb]
    \centering
    \scalebox{1}{
    \begin{tabular}{c|p{1.9cm}<{\centering}p{1.5cm}<{\centering}p{1.5cm}<{\centering}}
        \toprule
        Model & Entertainment & Readability & Relevance\\
        \midrule
        \multicolumn{4}{c}{penggen}\\
        \midrule
        \textbf{Gold} & 1.88 & 2.86 & 2.96 \\
        \textbf{BART} & 0.76 & 2.94 & 2.10\\
        \textbf{GPT} & 0.48 & 2.80 & 1.54\\
        \textbf{T5} & 0.90 & 2.92 & 1.98\\
        \textbf{Trans} & 0.38 & 2.94 & 1.16\\
        \midrule
        \multicolumn{4}{c}{dougen}\\
        \midrule
        \textbf{Gold} & 2.08 & 2.96 & 3.00\\
        \textbf{BART} & 0.82 & 2.94 & 2.22\\
        \textbf{GPT} & 0.64 & 3.00 & 1.86\\
        \textbf{T5} & 0.76 & 2.88 & 1.98\\
        \textbf{Trans} & 0.44 & 2.64 & 0.98\\
        \bottomrule
    \end{tabular}}
    \caption{Human evaluation of model outputs. They are rated on a scale from 0 to 3. \textbf{Gold} stands for the ground truth response.}
    \label{tab:human_evaluation}
\end{table} 
\begin{table*}[htb]
    \small
    \centering
    \begin{tabular}{c|m{2.5cm}<{\raggedright}p{8cm}}
        \toprule
        \midrule
         \multirow{5}{*}[-30pt]{Context} & 逗哏(dougen): & 孙悟空那肉，它塞牙。\newline
         (The meat of Monkey King will stuffed teeth.)
         \\
         ~ & 捧哏(penggen): & 太精瘦了。\newline
         (Too lean.)\\
         ~ & 逗哏(dougen): & 猪八戒，太糙。\newline
         (Zhu Bajie's meat is too tough.)\\
         ~ & 捧哏(penggen): & 是。\newline
         (That's true.)\\
         ~ & 逗哏(dougen): & 沙和尚，膻气。\newline
         (Sha Heshang's meat is too stinky).\\
        \midrule
        \multirow{1}{*}[-5pt]
        {Response} & 捧哏(penggen): & 可不。\newline
        (Sure enough.)\\
        \midrule
        \midrule
        \multirow{3}{*}[-30pt]{Context} & 逗哏(dougen): & 我们亲哥儿俩跟我师父练的功夫，师父都给我们起了名字。\newline(We brothers practiced kung fu with my master, who gave us both names.)\\
        ~ & 捧哏(penggen): & 都叫什么呢？\newline(What are the names?)\\
        ~ & 逗哏(dougen): & 我哥叫“白糖的”，我叫“澄沙馅儿的”。\newline(My brother's name is "white sugar" and I'm "red bean paste".)\\
        \midrule
        \multirow{1}{*}[-5pt]{Response} & 捧哏(penggen): & 澄沙的？\newline(Red bean paste?)\\
        \midrule
        \bottomrule
    \end{tabular}
    \caption{Sampled responses generated by baseline models. The first case shows one generic response which frequently appears in generated responses. The second case shows that the model responses with a question to express surprise, which is a typical skill in crosstalk performance.}
    \label{tab:case_study}
\end{table*} 
We employed two human annotators to assess 100 samples for the CRG task. Each generated response is assessed from three aspects. \textbf{Readability} measures the fluency of generated responses, including the grammar and phrase correctness; \textbf{Relevance} reflects the semantic relevance between context and response. It measures the logical and sentimental consistency of the dialogue as well; \textbf{Entertainment} reflects the level of humor of the response. For a better evaluation, we recruited two expert annotators from Tianjin, the origin of crosstalks, and familiar with the performing art of crosstalks.

Each annotator was presented with one context and five responses (including 1 golden response and 4 responses generated by different models), and asked to assign an integer score to each generated response with respect to each aspect. The scores are rated on a scale from 0(not at all) to 3(perfect without flaws). 

Results are shown in Table \ref{tab:human_evaluation}. Most models can gain comparable performance with golden responses in readability. However, it can clearly be seen that generated responses are quite poor in entertainment. It indicates a huge difference between the typical models' outputs and the real crosstalks since entertainment is the most important feature of crosstalks. We can also observe that pretrained models outperform models without pretraining in all aspects by a large margin. It again proves the usefulness of pretraining.

\subsection{Case Study}

We make case studies to better understand the performance of models for the CRG task. Some sampled cases are shown in Table \ref{tab:case_study}. We find that generated responses lack diversity, especially for models trained on \textbf{penggen}. Many generic replies appears frequently, such as ``这都不像话！(Nonsense!)”  or ``可不！(Sure enough!)” . At the same time, models are unable to start a new topic, no matter whether they are trained on \textbf{dougen} or \textbf{penggen}. However, we also notice that models have learned some simple language patterns in crosstalk performance such as rhetorical patterns.

Following the initial goal to automatically generate crosstalks, we paired the best model\footnote{According to the automated metrics, we used the T5 model in the experiment. } trained on \textbf{penggen} with the best one trained on \textbf{dougen}. Given the beginning of a human-written script, the two models were asked to play their corresponding roles, and generate responses alternately. All of the fifty generated crosstalks got stuck in repetitions of similar utterances after three rounds. The reason could be the lack of diversity and the similar patterns of generating responses. As a pipeline, this process also suffers from the accumulation of errors in every step.

\section{Conclusions and Future Work}

In this paper, we proposed the first open-source Chinese crosstalk dataset {\it CrossDial} for both the CRG task and the CRS task, and investigated the possibility of automatic generation of entertaining dialogues in Chinese crosstalks.  Through experiments, we found that special language patterns of Chinese crosstalks were difficult for current neural models. We provided two performance benchmarks and hoped that they would push forward the automatic generation of the traditional Chinese art.

In future work, we will try to exploit other dialogue data in similar domains to further improve the performance. We will also try to generate the whole crosstalk from scratch, which may be very beneficial for the crosstalk industry. Last but not least, we will explore automated metrics for a better evaluation of generated crosstalks from the perspective of humor.

\section{Limitations}
The crosstalk scripts in our dataset are from multiple sources. For audio-based transcriptions, error might be introduced. We have made efforts to clean the data yet still observed misspellings and incorrect punctuation in our final dataset.

Due to the cost constraints, we only recruited one annotator for each sample and annotated 100 data samples. We may perform a more thorough human evaluation for future works.

\section{Ethical Consideration}
We propose a novel dataset in this study. In terms of intellectual properties, all crosstalk scripts in our dataset are available on the Internet. 

We recruited annotators to evaluate our dataset. We compensated them for approximately ¥60 per hour. We first conducted in-house annotation to determine the speed. Typically, one hundred data samples took about one hour, and we compensated the workers for ¥0.6 per sample.

We manually reviewed the dataset and performed several rule-based methods to remove offensive language. Despite our efforts to minimize bias, there still can be some utterances that may trigger offenses.

\bibliography{anthology,custom}
\bibliographystyle{acl_natbib}

\end{CJK*}
\end{document}